\documentclass[letterpaper, 10 pt, conference]{ieeeconf}  

\IEEEoverridecommandlockouts                              

\usepackage{graphics} 
\usepackage{epsfig} 
\usepackage{mathptmx} 
\usepackage{times} 
\usepackage{amsmath} 
\usepackage{amssymb}  
\usepackage{colortbl}
\usepackage{tabularx, booktabs}
\newcolumntype{C}{>{\centering\arraybackslash}X} 

\usepackage{xspace}

\newcommand{\xrel}{x_{rel}}
\newcommand{\yrel}{y_{rel}}
\newcommand{\psirel}{\psi_{rel}}
\newcommand{\lr}{l_{r}}
\newcommand{\lf}{l_{f}}
\newcommand{\df}{\delta_{f}}

\newcommand{\reachpred}{Reachability-Pred\xspace}
\newcommand{\reachnopred}{Reachability-NoPred\xspace}

\title{\LARGE \bf
Prediction-Based Reachability for Collision Avoidance \\in Autonomous Driving}

\author{Anjian Li$^{1}$, Liting Sun$^{2}$, Wei Zhan$^{2}$, Masayoshi Tomizuka$^{2}$ and Mo Chen$^{1}$
\thanks{This work was supported by the NSERC Discovery Grant.}%
\thanks{$^{1}$Anjian Li and Mo Chen are with School of Computing Science, Simon Fraser University, Burnaby, BC, Canada, V5A 1S6
        {\tt\small \{anjianl, mochen\}@sfu.ca}}%
\thanks{$^{2}$Liting Sun, Wei Zhan and Masayoshi Tomizuka are with the Department of Mechanical Engineering, University of California,
Berkeley, CA 94720 USA
{\tt\small \{litingsun, wzhan,  tomizuka\}@berkeley.edu}}%
}

\renewcommand{\baselinestretch}{.92}

\begin{document}

\maketitle
\thispagestyle{empty}
\pagestyle{empty}

\begin{abstract}

Safety is an important topic in autonomous driving since any collision may cause serious injury to people and damage to property.
Hamilton-Jacobi (HJ) Reachability is a formal method that verifies safety in multi-agent interaction and provides a safety controller for collision avoidance.
However, due to the worst-case assumption on the car's future behaviours, reachability might result in too much conservatism such that the normal operation of the vehicle is badly hindered.
In this paper, we leverage the power of trajectory prediction and propose a prediction-based reachability framework to compute safety controllers. Instead of always assuming the worst case, we cluster the car's behaviors into multiple driving modes, e.g. left turn or right turn. Under each mode, a reachability-based safety controller is designed based on a less conservative action set.
For online implementation, we first utilize the trajectory prediction and our proposed mode classifier to predict the possible modes, and then deploy the corresponding safety controller.
Through simulations in a T-intersection and an 8-way roundabout, we demonstrate that our prediction-based reachability method largely avoids collision between two interacting cars and reduces the conservatism that the safety controller brings to the car's original operation.

\end{abstract}

\section{INTRODUCTION} \label{sec:intro}

With the surge of deep learning and advanced sensor technology, there has been great interest in developing autonomous agent that can perceive, analyze and predict the environment \cite{hancock2019future, grigorescu2020survey} and interact with humans. Ensuring safe control of such autonomous agents has always been a critical topic because collisions can lead to dramatic damage.

Various work has been done on collision avoidance for autonomous agents.
In traditional model-based optimal control, one often defines a large cost near the obstacles or adds hard/chance constraints when planning trajectories \cite{gu2015tunable}\cite{chen2017constrained}.
Recently, learning-based methods have also been adopted.
Autonomous agents can learn collision avoidance controllers from expert demonstrations via either imitation learning \cite{pan2018agile}\cite{li2020generating}\cite{sun2018fast} or inverse reinforcement learning \cite{kuderer2015learning}\cite{ sun2018probabilistic}\cite{wang2021socially}.
Model-free reinforcement learning has also been to learn safety maneuvers in complex and dynamic environment \cite{everett2019collision}\cite{cao2020reinforcement}.
However, most of the above learning-based methods cannot handle safety constraints. For instance, imitation learning cannot guarantee the safety of the generated actions, particularly for the end-to-end imitation models. 
Inverse reinforcement learning cannot accurately recover the reward functions in the presence of unknown safety constraints, and also hardly generalize well to real vehicles due to its sim-to-real pipeline. Moreover, for deep imitation learning and reinforcement learning, it is hard to analyze where the error comes from in the pipeline. Trajectories planned by finite-horizon decision and planning frameworks \cite{zhan2016ncds} can satisfy safety constraints within a finite horizon, but safety cannot be guaranteed for infinite horizon if no terminal constraints are included; researchers resort to reachable-set-based methods \cite{Pek2018iros} for such guarantees. 
\begin{figure}
    \vspace{4pt}
    \centering
    \includegraphics[width=1\linewidth]{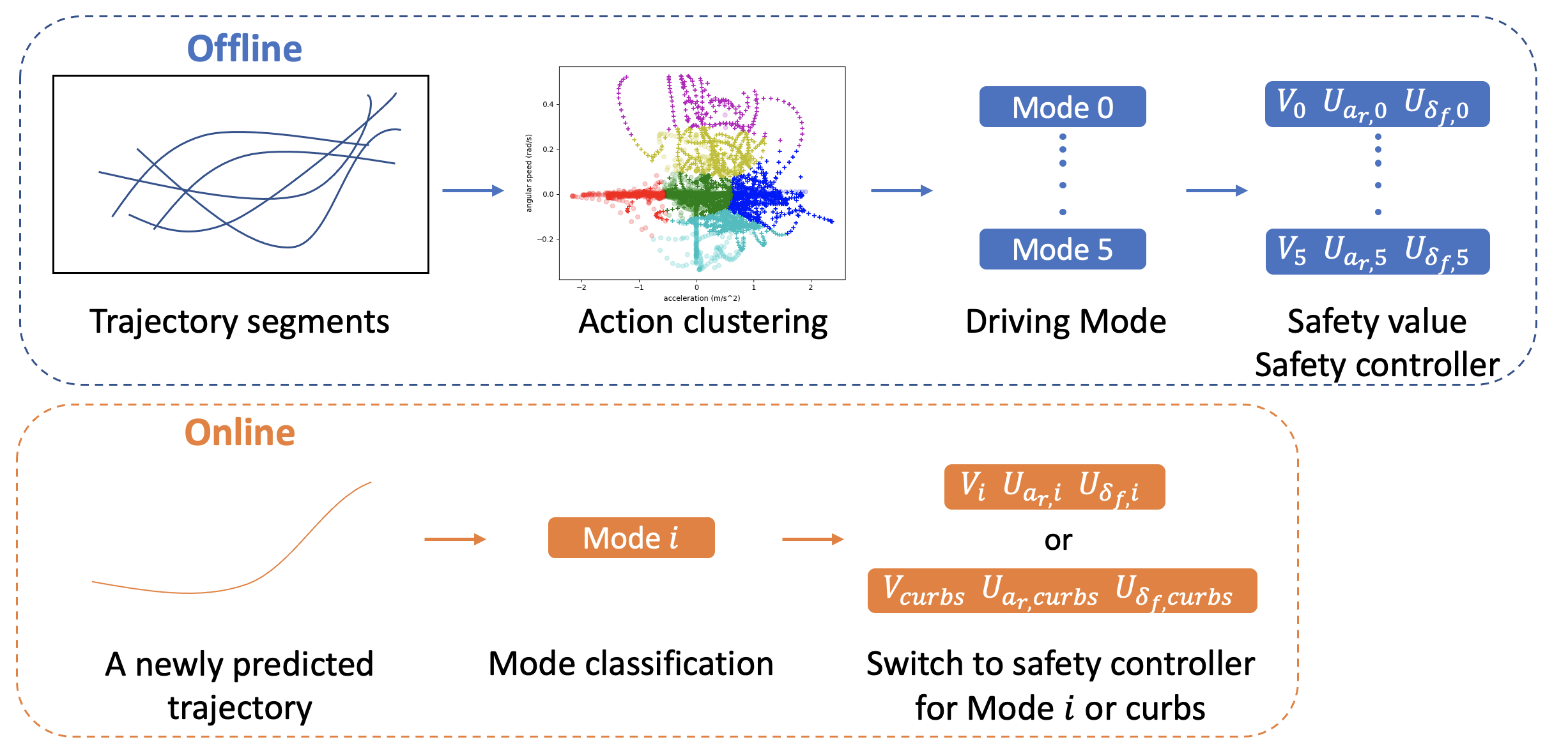}
    \caption{Work flow of our method. Offline, we cluster the trajectory into driving modes and compute BRTs (represented by safety value) and safety controllers as lookup tables for each mode.
    Online, as the robot car is running, whenever a newly predicted trajectory is given, it will be classified as certain driving mode.
    Then the robot car will check the safety value and switch to the safety controller for that mode or curbs when necessary.}
    \label{Fig:Workflow}
    \vspace{-1em}
\end{figure}

Hamilton-Jacobi (HJ) Reachability is a formal method that can verify the safety of the agents \cite{mitchell2005time}.
Given the agent's dynamics and the collision set of states as the target set, it computes Backward Reachable Tubes (BRT), and the agent will be guaranteed safe when staying outside of the BRT \cite{bansal2017hamilton, chen2018hamilton}.
This is achieved by assuming worst-case control and disturbance inputs to the dynamics, and by computing the global optimal solution in entire state space via dynamic programming \cite{mitchell2008flexible}.
For collision avoidance between two cars, relative dynamics can be used where we give control of the ``robot car" and consider the other ``human car" as disturbances in the dynamics \cite{merz1972game, mitchell2001games}.

Despite the advantages, HJ Reachability faces large challenges in autonomous driving.
First, traffic scenarios are often very crowded with intensive interactions between cars.
Since HJ Reachability always assumes the worst case of other cars' actions, the BRT can be too conservative so that the robot car can hardly operate normally.
Second, computing BRT suffers from curse of dimensionality.
Though approximating algorithms were developed \cite{chen2018decomposition, li2020guaranteed, herbert2019reachability}, it is still hard to compute BRTs for systems with 5D or higher-dimensional dynamics in real time.

There have been efforts focusing on less conservative and more practical BRTs.
In \cite{fisac2018general}, the authors designed an online learning framework to obtain more accurate bounds of wind disturbance.
In traffic scenarios, \cite{driggs2018robust} proposed the empirical reachable set, a probabilistic forward reachable set for cars, which rejected unlikely trajectories via non-parametric estimation.
The authors in \cite{leung2020infusing} modeled the interaction between the robot car and the human car as a pursuit evasion game and computed BRT for the relative system between the two cars, which is less conservative than considering forward reachable sets.
In this case, the BRT was precomputed for online use, but worst-case behaviors were assumed for human car and a projected safety controller was needed to avoid overly aggressive avoidance behaviors.

Motivated by above methods, we use trajectory prediction to reduce the action range of the human car, thereby obtaining more practical BRTs that preserve safe but are not overly conservative.
Luckily, with the development of deep learning, state-of-the-art trajectory prediction algorithms achieve great performance. For example, \cite{chai2019multipath} provided a probabilistic prediction of behaviors over candidate anchors, and \cite{rhinehart2019precog, zhao2020tnt} proposed goal-conditioned trajectory prediction networks.
\cite{hu2020scenario} proposed generic features that consider both dynamic and static information in the traffic and showed success in different scenarios.

\textbf{Contributions:} We aim to integrate general trajectory prediction into the design of reachability-based safety controllers to achieve more efficient two-car collision avoidance in real time with probabilistic safety guarantees.
First, we model the two-car interaction as pursuit evasion game, with the human car being the pursuer. 
Our additional key insight is that given the prediction of the human car's future trajectory, smaller action bounds can be used in the reachability computation, resulting in less conservative BRTs.
Second, a mode switch strategy is proposed to achieve real time BRT update.
Here, the intuition is that if the car is turning left, it is unlikely that it will suddenly turn right.
Thus, we cluster the human driver's behaviors into six common driving modes with associated action bounds.
The corresponding BRT for each mode is saved as a look-up table and switched online according to the prediction outcomes.
Our simulations show that our method not only preserves safety but also minimizes unnecessary impact to the car's original operations. 


\section{BACKGROUND} \label{sec:background}

\subsection{Hamilton-Jacobi Reachability} \label{subsec: HJ intro}

In this section, we introduce Hamilton-Jacobi (HJ) Reachability to verify safety in situations when a collision may happen between two cars.
In this two-car interaction, there is one car we take control of, named the robot car, and another car which we can only observe its actions, named the human car.
We assume that the dynamics for the robot car and the human car are defined respectively by the ODEs $\dot z_{r}(t) = f_{r}(z_{r}, u_{r})$ and $\dot z_{h}(t) = f_{h}(z_{h}, u_{h})$, where $t$ represents time, $z_{r} \in \mathbb{R}^{n_{r}}$ and $z_{h} \in \mathbb{R}^{n_{h}}$ are states, $u_{r} \in \mathcal U_{r}$, $u_{h} \in \mathcal U_{h}$ are controls for the robot and human car respectively.
We, the robot car, aim to use $u_{r}$ to avoid collision while considering a range of possible $u_{h}$ of the human car.

We model this situation as traditional pursuit-evasion game \cite{merz1972game}: the pursuer (human car) wants to catch and the evader (robot car) wants to avoid. 
Since collision depends only on how close the two cars are -- instead of the exact locations of the cars the collision happens -- we use relative dynamics $\dot z_{rel} = f_{rel}(z_{rel}, u_{rel}, d_{rel})$ to define the joint system where the relative states $z_{rel} \in \mathbb{R}^{n_{rel}}$ are constructed from $z_{r}$ and $z_{h}$.
The controls of the relative dynamics $u_{rel} \in \mathcal U_{rel}$ are the robot control $u_{r}$, and the disturbances $d_{rel} \in \mathcal D_{rel}$ are the human control $u_{h}$.
We define collision as the human car enters the target set $\mathcal{T}$ around the robot car.

In HJ Reachability, given the above relative dynamics and target set as the collision set, we compute Backward Reachable Tube (BRT), representing states that will inevitably enter the target set within some time horizon $T$ under worst disturbances $d_{rel}$ despite best controls $u_{rel}$ \cite{bansal2017hamilton}. 
We represent the target set $\mathcal{T}$ as the zero sub-level set of a signed distance function $l(z_{rel})$, where $z_{rel} \in \mathcal{T} \Leftrightarrow l(z_{rel}) \le 0$.
Then with the final value function $V(z_{rel},0) = l(z_{rel})$, we can solve the following HJ equation to obtain $V(z_{rel}, t)$ whose zero sub-level set represents the BRT \cite{chen2018hamilton}:
\begin{align} \label{Eq: HJ PDE}
    &\min \{ D_{t} V(z_{rel},t)+H(z_{rel}, \nabla V(z_{rel},t)), 
    V(z_{rel},0) - 
    V(z_{rel},t) \} \nonumber \\ &\qquad \qquad\qquad \qquad  = 0, \quad t \in [-T, 0] \nonumber \\ 
    &H(z_{rel}, \nabla V(z_{rel}, t)) = \nonumber \\
    &\qquad \min_{d_{rel} \in \mathcal D_{rel}} \max_{u_{rel} \in \mathcal U_{rel}}
    \nabla V(z_{rel},t) ^\top f_{rel}(z_{rel},u_{rel}, d_{rel})
\end{align}

The corresponding optimal safety controller $u_{rel}$ is:
\begin{align} \label{Eq: robot control}
        u_{rel}^{*}(t) {=} \arg \min_{d_{rel} \in \mathcal D_{rel}} \max_{u_{rel} \in \mathcal U_{rel}} \ \nabla V(z_{rel},t) ^\top f_{rel}(z_{rel},u_{rel}, d_{rel}).
\end{align}

By discretizing the state space into grids, \eqref{Eq: HJ PDE} can be solved via dynamic programming with level set method \cite{osher2004level, mitchell2005time}.
In this paper we use the optimzed\_dp toolbox \cite{optimizedp} with HeteroCL \cite{lai2019heterocl} to solve BRTs.

\subsection{Scenario-Transferable Probabilistic Prediction} \label{subsec: scenario_transferable}

To predict the car's future trajectory, we adopt the probabilistic prediction algorithm in \cite{hu2020scenario}.
Trajectory prediction in traffic is hard since
it needs to consider the interactions between cars and the road constraints, e.g. curbs.
The algorithm in \cite{hu2020scenario} is built on generic representation of both static and dynamic information of the environment, and is able to predict the future trajectory of a car of interest in highly interactive traffic and can transfer to different scenarios like intersections or roundabouts.

It should be noted that our method is adaptable to any trajectory prediction algorithm as long as it can predict series of future positions for cars.
However, the confidence of prediction will affect the probability of safety guarantee of our designed controller, discussed in Sec. \ref{Sec:mode switch}.

\section{DRIVING MODE ANALYSIS} \label{sec:driver_behaviour}

In this section, we aim to derive common human driving modes. For each mode, a unique action bound exists, so that less conservative BRT can be found in Sec. \ref{sec:reachability_based}.
We first collect data of predicted car trajectories with algorithm in \cite{hu2020scenario}, which is trained on the real world INTERACTION dataset \cite{interactiondataset}.
Then we cluster the trajectory segments into several driving modes, e.g. left turn or right turn, based on the linear acceleration and angular speed.
Finally we build a classifier to determine the probability distribution over each driving mode when a new predicted trajectory is given.
The procedure is summarized in the offline part in Fig. \ref{Fig:Workflow}.

\subsection{Trajectory Collecting and Processing}

For the selected car in traffic, \cite{hu2020scenario} predicts its $n$-step future trajectory $\{(x_{t}, y_{t}, v_{t})\}_{t=0}^{n \Delta t}$, where $x_{t}$, $y_{t}$ are the global $x$ and $y$ position, $v_{t}$ is the speed and the time interval $\Delta t = 0.1s$.
The predicted trajectories are collected from two different scenarios: T-intersection and 8-way roundabout.
Please refer to \cite{interactiondataset} for map details.


To obtain the human car's action at each time step, we use the extended Dubins Car to model the human car's dynamics $\dot z_{h} = f_{h}(z_{h}, u_{h})$:
\begin{align} \label{Eq: human car dynamics}
    \dot x_{h} &= v_{h} \cos \psi_{h}, &
    \dot y_{h} &= v_{h} \sin \psi_{h}  \nonumber \\
    \dot v_{h} &= a_{h}, &
    \dot \psi_{h} &= \omega_{h}
\end{align}
where the state $z_{h} = (x_{h}, y_{h}, v_{h}, \psi_{h})$ comprises the position $(x_h, y_h)$, linear speed $v_h$, and the orientation $\psi_{h} \in [-\pi, \pi)$.
The controls, or actions the human car can take, are the acceleration $a_{h} \in \mathcal{U}_{a_{h}}$ and angular speed $\omega_{h} \in \mathcal{U}_{\omega_{h}}$. 

Since \eqref{Eq: human car dynamics} is differentially flat, then the action dataset, $\{(a_{h, k}, \psi_{h, k})\}_{k=1}^{N}$ can be approximated from data in the form of $\{(x_{t}, y_{t}, v_{t})\}_{t=0}^{n \Delta t}$  \cite{walambe2016optimal}, where $N=2365$ for the T-intersection and $N=1911$ for the 8-way roundabout scenario.


\subsection{Driving Mode Clustering}
Common patterns and modes can be extracted \cite{jia2020ide} from large amounts of human driving data. In this paper, we use clustering to extract patterns of driving based on the acceleration $a_{h}$ and angular speed $\omega_{h}$.
Based on our knowledge of driving, before clustering, we pre-define six common driving modes in intersection and roundabout scenarios and set the nominal action $a_{h} (m/s)$ and $\omega_{h} (rad/s)$ for each mode:
\begin{align}
    &\text{Mode 0: Deceleration}, a_{h} = -1.5, \omega_{h} = 0 \nonumber \\
    &\text{Mode 1: Stable}, a_{h} = 0, \omega_{h} = 0 \nonumber \\
    &\text{Mode 2: Acceleration}, a_{h} = 1.5, \omega_{h} = 0 \nonumber \\
    &\text{Mode 3: Left turn}, a_{h} = 0, \omega_{h} = 0.2 \nonumber \\
    &\text{Mode 4: Right turn}, a_{h} = 0, \omega_{h} = - 0.25 \nonumber \\
    &\text{Mode 5: Roundabout}, a_{h} = 0, \omega_{h} = 0.4 \nonumber
\end{align}

\begin{figure}
    \vspace{5pt}
    \centering
    \includegraphics[width=0.95\linewidth]{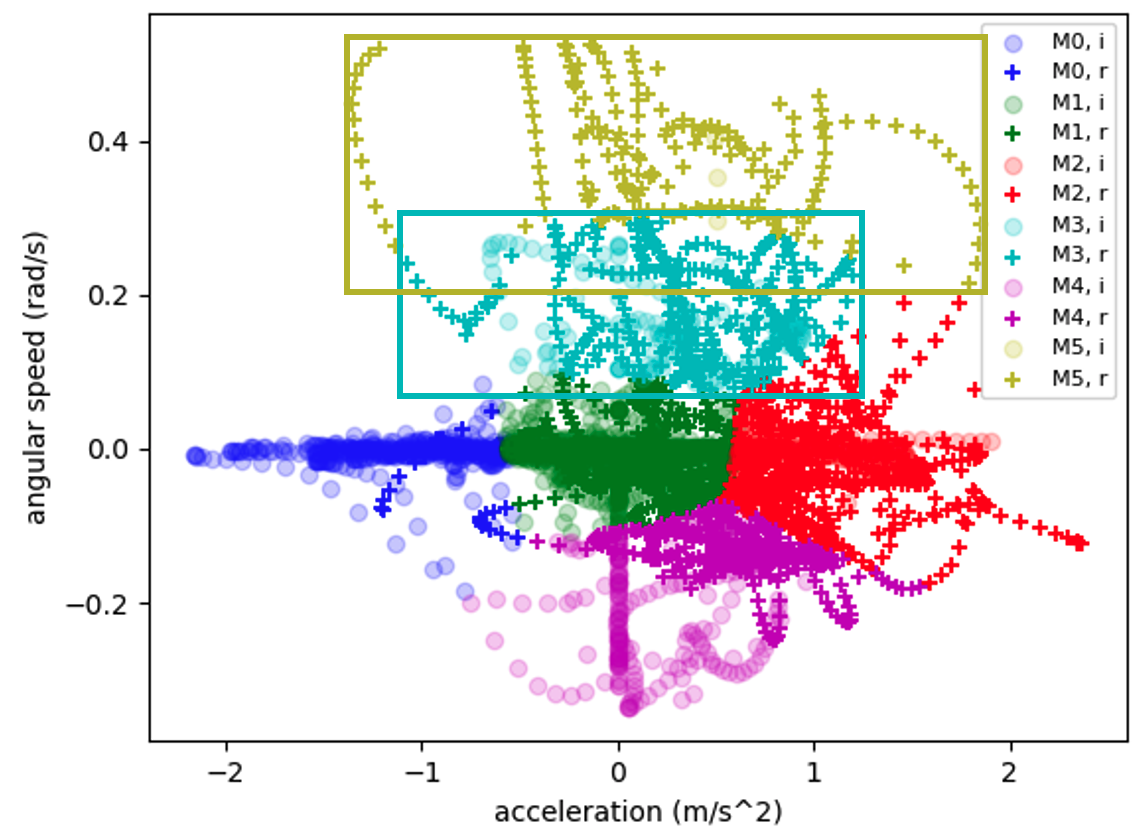}
    \caption{Driving mode clustering based on acceleration and angular speed. The entire dataset is clustered into 6 driving modes from Mode 0 to Mode 5. ``o" and ``+" represent data from intersection and roundabout scenario.
    The action range for each mode is the rectangle bounded by the outermost data points and may have some overlap. }
    \label{Fig:clustering}
    \vspace{-1em}
\end{figure}

For each $(a_{h, k}, \omega_{h,k})$ pair in the dataset, we first normalize them into $[-1, 1]$.
Then we construct a 6D clustering feature $(d_{M0}, d_{M1}, d_{M2}, d_{M3}, d_{M4}, d_{M5})$, each term being the datapoint's Euclidean distance to the six mode defaults.
With these features, we use k-means \cite{macqueen1967some} to cluster all the action data into 6 driving modes, shown in Fig. \ref{Fig:clustering}.

In Fig. \ref{Fig:clustering}, the $x$- and $y$-axis represent acceleration and angular speed respectively. ``Cross" and light ``dot" data points are from roundabout and intersection scenarios, and the whole action dataset is clustered nicely into six driving modes each centered around our mode default.
Each scenario has adequate data from Mode 0 to Mode 4, but only Mode 5 contains data mostly from roundabout scenario, because when cars are inside roundabout, they usually have a larger positive angular speed than a normal left turn.

For each mode, we define the corresponding action range $\mathcal{U}_{a_{h}} \times \mathcal{U}_{\omega_{h}}$ to be a rectangle bounded by the uppermost, lowermost, leftmost and rightmost data points.
For instance in Fig. \ref{Fig:clustering}, we consider any action pair $(a_{h}, \omega_{h})$ inside the cyan and yellow rectangle to be in Mode 3: Left turn and Mode 5: Roundabout, respectively.

\subsection{Probabilistic Mode Classifier} \label{sec:mode classifier}

Any pair $(a_{h}, \omega_{h})$ may fall into the action range of zero, one or more modes.
Thus given $(a_{h}, \omega_{h})$ we define its probability over each mode as follows: if it does not fall into a range of any mode, it will be regarded as ``Mode -1: Other";
if it only falls into the range of a single mode, it has 100\% probability to be in that mode and 0\% probability to be in other modes;
if it falls into the ranges of a set of modes $\{\text{Mode } i | i \in \sigma \}$, then the probability to be in Mode $j$, $j \in \sigma$, is $(1 / d_{j}) / (\sum_{i \in \sigma} 1 / d_{i})$, where $d_{i}$ is the distance to the closest rectangle boundary of Mode $i$.

\section{REACHABILITY-BASED SAFE CONTROLLER} \label{sec:reachability_based}

In this section we design the reachability-based safety controller for the robot car given the prediction of an observed human car.

\subsection{System Dynamics}

We use pursuit-evasion game in Sec. \ref{subsec: HJ intro} to model the pairwise interaction between two cars.
For the human car we define its dynamics as Eq. \eqref{Eq: human car dynamics}.
For our robot car we choose the higher-fidelity bicycle dynamics $\dot z_{r} = f_{r}(z_{r}, u_{r})$ \cite{kong2015kinematic}:
\begin{align} \label{Eq. robot car dynamics}
    \dot x_{r} &= v_{r} \cos (\psi_{r} + \beta_{r}) \nonumber \\
    \dot y_{r} &= v_{r} \sin (\psi_{r} + \beta_{r}) \nonumber \\
    \dot v_{r} &= a_{r} \nonumber \\
    \dot \psi_{r} &= \frac{v_{r}}{l_{r}} \sin (\beta_{r}) \nonumber \\
    \beta_{r} &= \tan^{-1} (\frac{l_{r}}{l_{f} + l_{r}} \tan (\delta_{f}))
\end{align}

\noindent where the state is $z_{r} = (x_{r}, y_{r}, v_{r}, \psi_{r})$. $(x_{r}, y_{r}) \in \mathbb{R}^2$ is global positions, $v_{r} \in \mathcal V_{r}$ is the linear speed, $\psi_{r} \in [-\pi, \pi)$ is the heading of the car, and $l_{f}, l_{r}$ are the distances from the center of mass to the front and rear axles respectively.
The control inputs are accelerations $a_{r} {\in} \mathcal U_{a_{r}}$ and steering angles $\df {\in} \mathcal U_{\df}$.
The control bounds are chosen so that the robot car has the same acceleration and turning ability as the human car.

Similar to \cite{leung2020infusing}, we define relative dynamics to be centered around the robot car and also consistent with its coordinate frame.
The relative $x$ position $x_{rel}$ is defined as the position in the robot car's orientation, and the relative y position $y_{rel}$ is perpendicular to the $\xrel$:
\begin{align} \label{Eq: coordinate transform}
    \begin{bmatrix}
    x_{rel}\\
    y_{rel}
    \end{bmatrix}
    =
    \begin{bmatrix}
    \cos \psi_{r} \quad \sin \psi_{r} \\
    -\sin \psi_{r} \quad \cos \psi_{r}
    \end{bmatrix}
    \begin{bmatrix}
    x_{h} - x_{r}\\
    y_{h} - y_{r}
    \end{bmatrix}
\end{align}

The relative angle $\psirel$ is defined based on the robot car's orientation $\psirel := \psi_{h} - \psi_{r}$.
However, the speed of the two cars are in their own coordinate frame, so we include both of them individually.
Finally we have the following 5D relative dynamics $\dot z_{rel} = f_{rel}(z_{rel}, u_{rel}, d_{rel})$:
\begin{align} \label{Eq: 5d relative dynamics}
    \dot x_{rel} &= \frac{v_{r}}{\lr} \sin (\beta_{r})  y_{rel} + v_{h} \cos \psi_{rel} - v_{r}  \cos \beta_{r} \nonumber \\
    \dot y_{rel} &= - \frac{v_{r}}{\lr} \sin (\beta_{r}) x_{rel} + v_{h}  \sin \psi_{rel} - v_{r}  \sin \beta_{r} \nonumber \\
    \dot \psi_{rel} &= \omega_{h} - \frac{v_{r}}{\lr} \sin (\beta_{r}) \nonumber \\
    \dot v_{h} &= a_{h} \nonumber \\
    \dot v_{r} &= a_{r} \nonumber \\
    \beta_{r} &= \arctan(\frac{\lr}{\lf + \lr} \tan (\df))
\end{align}

The state is $z_{rel} = (\xrel, \yrel, \psirel, v_{h}, v_{r})$. The control inputs are the robot car's controls $a_{r} \in \mathcal U_{a_{r}}$ and $\df \in \mathcal U_{\df}$.
The human car's controls $a_{h} \in \mathcal D_{a_{h}}$ and $\omega_{h} \in \mathcal D_{\omega_{h}}$ are considered as disturbances since the robot car does not have the ability to choose them.
Here $\mathcal D_{a_{h}}$ and $\mathcal D_{\omega_{h}}$ are equal to $\mathcal U_{a_{h}}$ and $\mathcal U_{\omega_{h}}$ in Sec. \ref{sec:driver_behaviour}, respectively.

\subsection{Collision Avoidance Between Cars} \label{subsec: BRT in different mode}

Given the relative dynamics, we set the target set $\mathcal{T}$ to be the collision set which is a rectangle centered around the robot car $\mathcal{T} := \{z_{rel} \mid |\xrel| \leq C_{1}, |\yrel| \leq C_{2} \}$.
The corresponding infinite time horizon BRT represents the states from which collision between the human car and the robot car is inevitable.
In this paper, to approximate the infinite time horizon BRT, we compute Eq. \eqref{Eq: HJ PDE} for a sufficient time horizon until the value function $V$ is converged.

One of our key contributions is to have less conservative BRTs, and it is achieved by having a 
smaller range of disturbance set $\mathcal D_{a_{h}}$ and $\mathcal D_{\omega_{h}}$.
Recall that in Sec. \ref{sec:driver_behaviour}, we summarized six common driving modes with different $\mathcal D_{a_{h}}$ and $\mathcal D_{\omega_{h}}$ from clustering, thus we solve Eq. \eqref{Eq: HJ PDE} for each mode individually and obtain the safety value $V_{i}(z_{rel})$ for Mode $i$, whose zero sub-level set is the BRT.
In addition, we compute another $V_{-1}(z_{rel})$ and BRT for Mode $-1$ using the physical limit of the car, i.e. assuming the worst-case behavior from the human car.

\begin{figure}
    \vspace{4pt}
    \centering
    \includegraphics[width=0.95\linewidth]{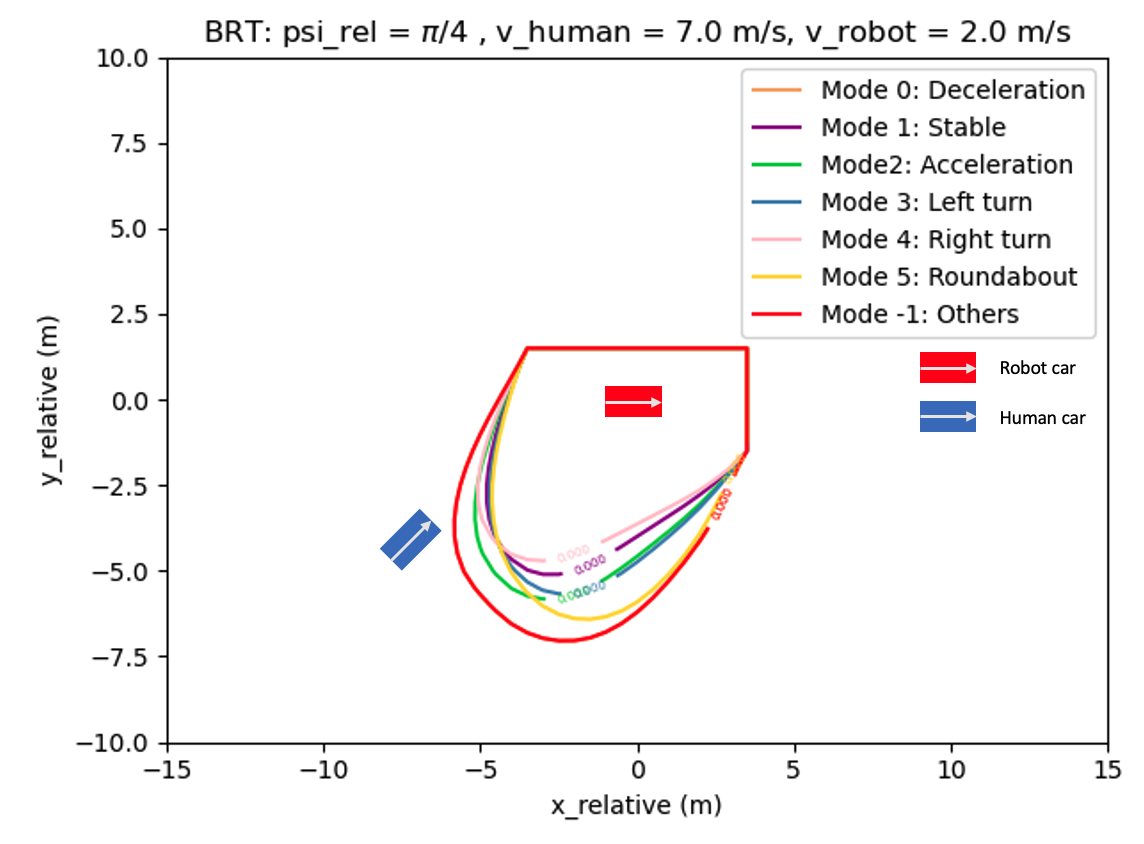}
    \caption{2D slice of BRT for different driving modes. Mode $-1$ assumes full range of human car's action limit and thus has the largest BRT.
    From Mode $0$ to Mode $5$ the BRTs are less conservative in different direction.
    }
    \label{Fig:reachable set}
    \vspace{-1em}
\end{figure}

In Fig. \ref{Fig:reachable set}, we show the comparison of BRTs for different driving modes.
Since the full BRT is 5D, we show the 2D slice at $\psirel = \pi/4, v_{h} = 6m/s, v_{r} = 1m/s$.
As seen in Fig. \ref{Fig:reachable set}, the BRT of Mode $-1$ is the largest since it considers all possible controls of the human car as long as within the car's physical limit, so it is reasonable that the robot car has to be further away to maintain safety.
If we believe that the human car is in some driving mode, for example in Mode 3: Left turn, its acceleration and angular speed will be restricted, and as a result, the BRT is smaller and sways more to the right side.

For each Mode $i$, besides the safety value $V_{i}(z_{rel})$, we also compute Eq. \eqref{Eq: robot control} to obtain the robot car's safety control $a_{r}$ and $\df$ and save them as lookup tables $U_{a_{r}, i}(z_{rel})$ and $U_{\df, i}(z_{rel})$.
When the human car is in Mode $i$, our robot car will only track the safety value $V_{i}(z_{rel})$ and consider using safe controller $U_{a_{r}, i}(z_{rel})$ and $U_{\df, i}(z_{rel})$.

\subsection{Obstacle Avoidance for Curbs} \label{sec:curb controller}

In real traffic scenarios, cars not only have to avoid the traffic in the lane, but also need to avoid collision with the curbs.
Therefore to design a more practical safety controller, we incorporate another safety controller for curbs and other static obstacles only.

Following the traditional reachability setting in \cite{chen2018hamilton}, we set the target set to be the curb area obtained from the INTERACTION dataset \cite{hu2020scenario}, and use the dynamics in Eq. \eqref{Eq. robot car dynamics} to describe the robot car.
By solving HJ equation to convergence, we can approximate the infinite time horizon BRT, and save the safety value $V_{curbs}(z_{r})$ and safety controller $U_{a_{r}, curbs}(z_{r})$ and $U_{\df, curbs}(z_{r})$ as lookup tables.

\subsection{Online Mode Switch Strategy} \label{Sec:mode switch}

Usually it is very hard to keep an up-to-date BRT for safety checks when the car is operating online, because the BRT for 5D dynamics takes hours to compute.
But with the precomputed safety values and safety controllers in Sec. \ref{subsec: BRT in different mode} and \ref{sec:curb controller}, we can achieve real-time safe control by switching to the appropriate look-up table based on predictions, which takes trivial time.

We use a mode switch strategy shown as the online part in Fig. \ref{Fig:Workflow}.
Here, accurate state estimation is assumed for both cars.
The robot car continuously observes the human car's behaviors to update the prediction of the human car's future trajectory with algorithm in Sec. \ref{subsec: scenario_transferable}.
Every time the prediction is updated, we will infer the driving mode as described in Sec. \ref{sec:mode classifier}.
Suppose the human car is in Mode $i$ and its original controllers are $\Tilde a_{r}$ and $\Tilde \df$, we design hybrid controllers $\hat a_{r}$ and $\hat \df$ for the robot car where the safety controller may take over as follows:
\begin{align} \label{Eq:Hybrid controller designs}
    &\text{If } \min (V_{i}(z_{rel}), V_{curbs}(z_{r})) > 0, \text{ use original controllers} \nonumber \\
    &\hat a_{r} = \Tilde a_{r}, \hat \df = \Tilde \df ; \nonumber \\
    &\text{Else if } V_{i}(z_{rel}) \leq V_{curbs}(z_{r}), \text{ use safety controller for cars} \nonumber \\  
    &\hat a_{r} = U_{a_{r}, i}(z_{rel}), \hat \df = U_{\df, i}(z_{rel}) \nonumber ; \\
    &\text{Else, use safety controller for curbs} \nonumber \\
    &\hat a_{r} = U_{a_{r}, curbs}(z_{r}), \hat \df = U_{\df, curbs}(z_{r}).
\end{align}

The intuition is that, when the car is far from both curbs and other cars, it operates as normal.
Otherwise, whether it chooses safety controllers for curbs or for cars depends on which is the closest to colliding with the robot car.

Under this strategy, the safety guarantee is preserved in a probabilistic way.
Let $p_{predict}$ be the probability of the predicted trajectory from \cite{hu2020scenario}, and let $p_{mode}$ be the probability of this trajectory being in certain driving mode from Sec. \ref{sec:mode classifier}, and our designed safety controller in Eq. \eqref{Eq:Hybrid controller designs} can guarantee the safety of the human car and robot car with the probability $p_{safety} = p_{predict} \times p_{mode}$,
with perfect modeling and state estimation assumed.

\section{SIMULATION} \label{sec:simiulation}
In this section, we simulate the situation where a controlled robot car and a human car interact in two traffic scenarios.
When their planned paths have some overlap, without any safety controller, collision may happen in various ways.
We demonstrate that with our proposed prediction-based safety controller, the collision is largely avoided while unnecessary impact to the robot car is limited.
In comparison, our baseline uses reachability-based safety controller without any prediction.
Although safety is also preserved with the baseline method, the robot car deviates more from its originally planned path due to unnecessary and conservative override of the safety controller.
Note that our reachability formulation does not depend on any specific traffic scenario; the advantages of our method can also generalize across scenarios of intersection and roundabout.
\begin{table}[ht]
\vspace{0.5em}
\caption{Number of trials where the minimum distance of the two car is less than or equal to 0.5m/1m.} 
\centering 
\begin{tabular}{c| rr | rr | rr } 
\hline\hline 
\textbf{Intersection}
 &\multicolumn{2}{c}{Case 1} & \multicolumn{2}{c}{Case 2} & \multicolumn{2}{c}{Case 3} \\  
Method &$\leq0.5$ & $\leq1$ &$\leq0.5$ & $\leq1$ &$\leq0.5$ & $\leq1$ \\ [0.25ex] 
\hline   
Default & 3 & 10 & 1 & 3 & 10 & 11 \\  
\reachpred & \textbf{0} & \textbf{0} & \textbf{0} & \textbf{0} & \textbf{1} & 3 \\ 
\reachnopred  & \textbf{0} & \textbf{0} & \textbf{0} & \textbf{0} & \textbf{1} & \textbf{2} \\  
\hline\hline
\textbf{Roundabout}
 &\multicolumn{2}{c}{Case 1} & \multicolumn{2}{c}{Case 2} & \multicolumn{2}{c}{Case 3} \\  
Method &$\leq0.5$ & $\leq1$ &$\leq0.5$ & $\leq1$ &$\leq0.5$ & $\leq1$ \\ [0.25ex] 
\hline   
Default & 7 & 15 & 3 & 6 & 8 & 10 \\  
\reachpred & 1 & 1 & \textbf{0} & \textbf{0} & \textbf{0} & \textbf{0} \\ 
\reachnopred  & \textbf{0} & \textbf{0} &\textbf{0} & \textbf{0} & \textbf{0} & \textbf{0} \\  
\hline\hline 
\end{tabular} 
\label{table: collision rate} 
\vspace{-1em}
\end{table} 

\begin{table}[ht] 
\vspace{7pt}
\caption{Average and Maximum Deviation (m) over each trial.} 
\centering 
\begin{tabular}{c| rr | rr | rr } 
\hline\hline 
\textbf{Intersection}
 &\multicolumn{2}{c}{Case 1} & \multicolumn{2}{c}{Case 2} & \multicolumn{2}{c}{Case 3} \\  
Method &avg. & max &avg. & max &avg. & max \\ [0.25ex] 
\hline   
\reachpred & 3.39 & 16.70 & \textbf{1.78} & \textbf{3.32} & \textbf{1.45} & \textbf{2.98} \\ 
\reachnopred  & \textbf{2.01} & \textbf{4.02} & 2.10 & 5.81 & 1.82 & 3.53 \\  
\hline\hline  
\textbf{Roundabout}
 &\multicolumn{2}{c}{Case 1} & \multicolumn{2}{c}{Case 2} & \multicolumn{2}{c}{Case 3} \\  
Method &avg. & max &avg. & max &avg. & max \\ [0.25ex] 
\hline   
\reachpred & \textbf{2.32} & \textbf{3.96} & \textbf{2.95} & \textbf{7.17} & \textbf{1.90} & \textbf{4.33} \\ 
\reachnopred  & 2.81 & 4.60 & 4.70 & 16.62 & 2.36 & 5.05 \\  
\hline\hline  
\end{tabular} 
\label{table: deviation} 
\vspace{-1em}
\end{table}

\begin{figure*}
    \vspace{5pt}
    \centering
    \includegraphics[width=1\linewidth]{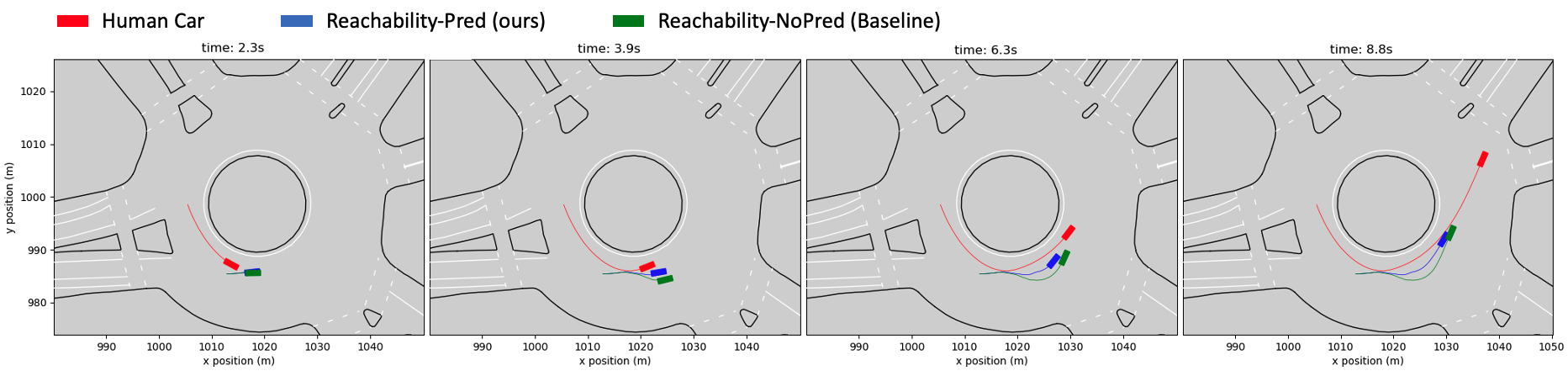}
    \caption{The trajectory comparison between \reachpred and \reachnopred when meeting a dangerous human car in roundabout scenario.
    The \reachnopred deviates more to maintain safety than \reachpred.}
    \label{Fig:trajectory}
    \vspace{-1em}
\end{figure*}

\subsection{Simulation Details}

\subsubsection{Method}
We compare three different safety controllers for the robot car.
The first is our proposed prediction-based safety controller using reachability, called \reachpred, which updates the BRT along with the latest prediction of the human car's driving mode.
The second is our baseline method which uses traditional reachability safety controller without prediction, called \reachnopred.
It keeps using the same BRT online considering all possible actions of the human car.
The third method is the default controller where no safety controller is involved.

\subsubsection{Path planning}

We select a T-intersection and an 8-way roundabout scenario from INTERACTION dataset \cite{interactiondataset}.
Our robot car follows a reference path which a real car has taken in the dataset.
With the adopted Stanley steering control \cite{thrun2006stanley} and PID speed control, the robot car tracks the reference path with a constant target speed of $2$ m/s.
We also want the human car to imitate a road user's behaviors.
Since we need to predict the human car's future trajectories, to simplify, we just let the human car operate exactly like the prediction output which is very close to a real car trajectory in the dataset.
In this case, the trajectory prediction of the human car is assumed to be 100\% correct.

\subsubsection{Test case}

For each scenario, we simulate 3 cases with different reference paths the robot and human car may take.
In each case, we run 10 or 20 trials with different start positions for the robot car.
The setting allows us to test how the safety controller reacts when the robot car meets the human car at its front, middle or the back side.

\begin{figure}[h]
    \centering
   \includegraphics[width=0.95\linewidth]{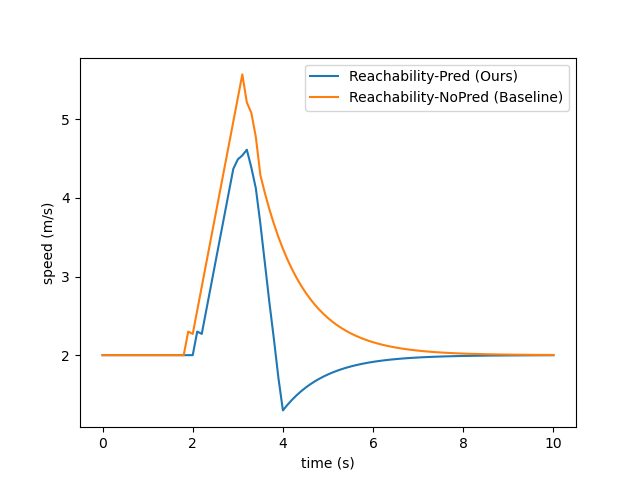}
    \caption{The speed profiles comparison for \reachpred and \reachnopred in roundabout scenario.
    For \reachnopred, the safety controller takes over earlier than \reachpred and pushes the robot car to a more extreme speed.}
    \label{Fig:speed profile}
    \vspace{-1em}
\end{figure}

\begin{table}[ht] 
\caption{Safety controller time for avoiding cars and curbs.} 
\centering 
\begin{tabular}{c| rr | rr | rr } 
\hline\hline 
\textbf{Intersection}
 &\multicolumn{2}{c}{Case 1} & \multicolumn{2}{c}{Case 2} & \multicolumn{2}{c}{Case 3} \\  
Method & car & curb & car & curb & car & curb \\ [0.25ex] 
\hline   
\reachpred & \textbf{7.81} & 9.26 & \textbf{7.81} & \textbf{0.00} & \textbf{13.76} & \textbf{2.95} \\ 
\reachnopred  & 7.95 & \textbf{1.76} & \textbf{7.81} & \textbf{0.00} & 17.10 & 5.52 \\  
\hline\hline  
\textbf{Roundabout}
 &\multicolumn{2}{c}{Case 1} & \multicolumn{2}{c}{Case 2} & \multicolumn{2}{c}{Case 3} \\  
Method & car & curb & car & curb & car & curb \\ [0.25ex] 
\hline   
\reachpred & \textbf{11.45} & \textbf{0.00} & \textbf{11.19} & 1.33 & \textbf{10.10} & \textbf{0.00} \\ 
\reachnopred  & 12.95 & \textbf{0.00} & 17.90 & \textbf{1.19} & 11.14 & \textbf{0.00} \\  
\hline\hline  
\end{tabular} 
\label{table: control time} 
\vspace{-1em}
\end{table}

\subsection{Simulation Result}

\subsubsection{Safety controller vs. no safety controller}

First, we demonstrate how the safety controller helps collision avoidance. We summarize the statistics of minimum distance between two cars over all trials.
In Table \ref{table: collision rate} we count the number of time steps that the two cars are closer than 0.5m/1m in each case.
In all case without safety controller, there are several trials that two cars are too close to each other and may cause collision.
With \reachnopred and \reachpred, the number of collisions is significantly reduced.
Note that, although \reachpred takes a much less conservative way for safety controller since it only considers a subset of human car's actions, it is almost as good as \reachnopred for collision avoidance when a perfect prediction is given.

\subsubsection{Prediction vs. No prediction}

Besides preserving safety of the car, our proposed \reachpred enables smoother operation of the robot car with less impact from the safety controller compared to \reachnopred.
We verify this by computing the average deviation and maximum deviation from the planned path, and the time that the safety controller takes effect in each trial.
We can see from Table \ref{table: deviation} and Table \ref{table: control time} that, in every case besides case 1 in intersection, \reachnopred makes the car deviate more from its path to maintain safety. The safety controller is activated more often and longer, which lowers the efficiency of the robot car for achieving its own goal.

Fig.~\ref{Fig:trajectory} and Fig. \ref{Fig:speed profile} show, respectively, the trajectories and the speed profiles of our \reachpred (blue) and \reachnopred (green) when they meet the human car (red).
In Fig.~\ref{Fig:trajectory} it is clear that without prediction the robot car will deviate more from the path.
In Fig.~\ref{Fig:speed profile} we find that the safety controller of \reachnopred starts earlier than \reachpred and leads the car to a more extreme speed.

Furthermore, in reality there might be road users other than our two-car system. Thus large deviation will hinder the traffic and expose the robot car to higher risk.

\subsubsection{Generalizability across scenarios}

Our method demonstrates better collision rate, smaller deviation and less interruption by the safety controller in both intersection and roundabout scenarios, based on Table \ref{table: collision rate}, \ref{table: deviation} and \ref{table: control time}.

\subsection{Error Case Analysis}

We check the failure case of our designed safety controller, i.e., case 1 in intersection. The robot car first uses the safety controller to avoid the oncoming human car, which leads it to the boundary of the upper curb.
Afterward, the safety controller for the curbs takes effect and forces the car to keep going up, which further increases the deviation.
Finally the car loses the ability to track its own reference path.
To solve this, we need to have a planner with a better higher-level decision-making system, which is out of this paper's scope.

\section{CONCLUSIONS} \label{sec:conclusion}

In this paper we present a prediction-based safety controller for two-car collision avoidance using HJ Reachability.
For each clustered driving mode, less conservative BRT and safety controller are precomputed and then switched online when the prediction of the human car is updated.
Simulation shows our work is superior in bringing less impact to the car's original operation while maintaining safety.

We believe this is a step forward to make HJ Reachability-based safety controller more practical in crowded scenarios for autonomous cars and ground robots.
For future work, we hope to extend our method to multi-agent collision avoidance, and want to incorporate it into vision-based perception and planning in partially observed environment.








\bibliographystyle{IEEEtran}
\bibliography{bibliography}

\begin{thebibliography}{10}
\providecommand{\url}[1]{#1}
\csname url@samestyle\endcsname
\providecommand{\newblock}{\relax}
\providecommand{\bibinfo}[2]{#2}
\providecommand{\BIBentrySTDinterwordspacing}{\spaceskip=0pt\relax}
\providecommand{\BIBentryALTinterwordstretchfactor}{4}
\providecommand{\BIBentryALTinterwordspacing}{\spaceskip=\fontdimen2\font plus
\BIBentryALTinterwordstretchfactor\fontdimen3\font minus
  \fontdimen4\font\relax}
\providecommand{\BIBforeignlanguage}[2]{{%
\expandafter\ifx\csname l@#1\endcsname\relax
\typeout{** WARNING: IEEEtran.bst: No hyphenation pattern has been}%
\typeout{** loaded for the language `#1'. Using the pattern for}%
\typeout{** the default language instead.}%
\else
\language=\csname l@#1\endcsname
\fi
#2}}
\providecommand{\BIBdecl}{\relax}
\BIBdecl

\bibitem{hancock2019future}
P.~A. Hancock, I.~Nourbakhsh, and J.~Stewart, ``On the future of transportation
  in an era of automated and autonomous vehicles,'' \emph{Proceedings of the
  National Academy of Sciences}, vol. 116, no.~16, pp. 7684--7691, 2019.

\bibitem{grigorescu2020survey}
S.~Grigorescu, B.~Trasnea, T.~Cocias, and G.~Macesanu, ``A survey of deep
  learning techniques for autonomous driving,'' \emph{Journal of Field
  Robotics}, vol.~37, no.~3, pp. 362--386, 2020.

\bibitem{gu2015tunable}
T.~Gu, J.~Atwood, C.~Dong, J.~M. Dolan, and J.-W. Lee, ``Tunable and stable
  real-time trajectory planning for urban autonomous driving,'' in \emph{2015
  IEEE/RSJ International Conference on Intelligent Robots and Systems
  (IROS)}.\hskip 1em plus 0.5em minus 0.4em\relax IEEE, 2015, pp. 250--256.

\bibitem{chen2017constrained}
J.~Chen, W.~Zhan, and M.~Tomizuka, ``Constrained iterative lqr for on-road
  autonomous driving motion planning,'' in \emph{2017 IEEE 20th International
  Conference on Intelligent Transportation Systems (ITSC)}.\hskip 1em plus
  0.5em minus 0.4em\relax IEEE, 2017, pp. 1--7.

\bibitem{pan2018agile}
Y.~Pan, C.-A. Cheng, K.~Saigol, K.~Lee, X.~Yan, E.~Theodorou, and B.~Boots,
  ``Agile autonomous driving using end-to-end deep imitation learning,'' in
  \emph{Robotics: science and systems}, 2018.

\bibitem{li2020generating}
A.~Li, S.~Bansal, G.~Giovanis, V.~Tolani, C.~Tomlin, and M.~Chen, ``Generating
  robust supervision for learning-based visual navigation using hamilton-jacobi
  reachability,'' in \emph{Learning for Dynamics and Control}, 2020, pp.
  500--510.

\bibitem{sun2018fast}
L.~Sun, C.~Peng, W.~Zhan, and M.~Tomizuka, ``A fast integrated planning and
  control framework for autonomous driving via imitation learning,'' in
  \emph{Dynamic Systems and Control Conference}, vol. 51913.\hskip 1em plus
  0.5em minus 0.4em\relax American Society of Mechanical Engineers, 2018.

\bibitem{kuderer2015learning}
M.~Kuderer, S.~Gulati, and W.~Burgard, ``Learning driving styles for autonomous
  vehicles from demonstration,'' in \emph{2015 IEEE International Conference on
  Robotics and Automation (ICRA)}.\hskip 1em plus 0.5em minus 0.4em\relax IEEE,
  2015, pp. 2641--2646.

\bibitem{sun2018probabilistic}
L.~Sun, W.~Zhan, and M.~Tomizuka, ``Probabilistic prediction of interactive
  driving behavior via hierarchical inverse reinforcement learning,'' in
  \emph{2018 21st International Conference on Intelligent Transportation
  Systems (ITSC)}.\hskip 1em plus 0.5em minus 0.4em\relax IEEE, 2018, pp.
  2111--2117.

\bibitem{wang2021socially}
L.~Wang, L.~Sun, M.~Tomizuka, and W.~Zhan, ``Socially-compatible behavior
  design of autonomous vehicles with verification on real human data,''
  \emph{IEEE Robotics and Automation Letters}, 2021.

\bibitem{everett2019collision}
M.~Everett, Y.~F. Chen, and J.~P. How, ``Collision avoidance in pedestrian-rich
  environments with deep reinforcement learning,'' \emph{arXiv preprint
  arXiv:1910.11689}, 2019.

\bibitem{cao2020reinforcement}
Z.~Cao, E.~B{\i}y{\i}k, W.~Z. Wang, A.~Raventos, A.~Gaidon, G.~Rosman, and
  D.~Sadigh, ``Reinforcement learning based control of imitative policies for
  near-accident driving,'' \emph{arXiv preprint arXiv:2007.00178}, 2020.

\bibitem{zhan2016ncds}
W.~{Zhan}, C.~{Liu}, C.~{Chan}, and M.~{Tomizuka}, ``A non-conservatively
  defensive strategy for urban autonomous driving,'' in \emph{2016 IEEE 19th
  International Conference on Intelligent Transportation Systems (ITSC)}, 2016,
  pp. 459--464.

\bibitem{Pek2018iros}
C.~{Pek} and M.~{Althoff}, ``Efficient computation of invariably safe states
  for motion planning of self-driving vehicles,'' in \emph{2018 IEEE/RSJ
  International Conference on Intelligent Robots and Systems (IROS)}, 2018, pp.
  3523--3530.

\bibitem{mitchell2005time}
I.~M. Mitchell, A.~M. Bayen, and C.~J. Tomlin, ``A time-dependent
  hamilton-jacobi formulation of reachable sets for continuous dynamic games,''
  \emph{IEEE Transactions on automatic control}, vol.~50, no.~7, pp. 947--957,
  2005.

\bibitem{bansal2017hamilton}
S.~Bansal, M.~Chen, S.~Herbert, and C.~J. Tomlin, ``Hamilton-jacobi
  reachability: A brief overview and recent advances,'' in \emph{2017 IEEE 56th
  Annual Conference on Decision and Control (CDC)}.\hskip 1em plus 0.5em minus
  0.4em\relax IEEE, 2017, pp. 2242--2253.

\bibitem{chen2018hamilton}
M.~Chen and C.~J. Tomlin, ``Hamilton--jacobi reachability: Some recent
  theoretical advances and applications in unmanned airspace management,''
  \emph{Annual Review of Control, Robotics, and Autonomous Systems}, vol.~1,
  pp. 333--358, 2018.

\bibitem{mitchell2008flexible}
I.~M. Mitchell, ``The flexible, extensible and efficient toolbox of level set
  methods,'' \emph{Journal of Scientific Computing}, vol.~35, no. 2-3, pp.
  300--329, 2008.

\bibitem{merz1972game}
A.~Merz, ``The game of two identical cars,'' \emph{Journal of Optimization
  Theory and Applications}, vol.~9, no.~5, pp. 324--343, 1972.

\bibitem{mitchell2001games}
I.~Mitchell, ``Games of two identical vehicles,'' Citeseer, Tech. Rep., 2001.

\bibitem{chen2018decomposition}
M.~Chen, S.~L. Herbert, M.~S. Vashishtha, S.~Bansal, and C.~J. Tomlin,
  ``Decomposition of reachable sets and tubes for a class of nonlinear
  systems,'' \emph{IEEE Transactions on Automatic Control}, vol.~63, no.~11,
  pp. 3675--3688, 2018.

\bibitem{li2020guaranteed}
A.~Li and M.~Chen, ``Guaranteed-safe approximate reachability via state
  dependency-based decomposition,'' in \emph{2020 American Control Conference
  (ACC)}.\hskip 1em plus 0.5em minus 0.4em\relax IEEE, 2020, pp. 974--980.

\bibitem{herbert2019reachability}
S.~L. Herbert, S.~Bansal, S.~Ghosh, and C.~J. Tomlin, ``Reachability-based
  safety guarantees using efficient initializations,'' in \emph{2019 IEEE 58th
  Conference on Decision and Control (CDC)}.\hskip 1em plus 0.5em minus
  0.4em\relax IEEE, 2019, pp. 4810--4816.

\bibitem{fisac2018general}
J.~F. Fisac, A.~K. Akametalu, M.~N. Zeilinger, S.~Kaynama, J.~Gillula, and
  C.~J. Tomlin, ``A general safety framework for learning-based control in
  uncertain robotic systems,'' \emph{IEEE Transactions on Automatic Control},
  vol.~64, no.~7, pp. 2737--2752, 2018.

\bibitem{driggs2018robust}
K.~Driggs-Campbell, R.~Dong, and R.~Bajcsy, ``Robust, informative
  human-in-the-loop predictions via empirical reachable sets,'' \emph{IEEE
  Transactions on Intelligent Vehicles}, vol.~3, no.~3, pp. 300--309, 2018.

\bibitem{leung2020infusing}
K.~Leung, E.~Schmerling, M.~Zhang, M.~Chen, J.~Talbot, J.~C. Gerdes, and
  M.~Pavone, ``On infusing reachability-based safety assurance within planning
  frameworks for human--robot vehicle interactions,'' \emph{The International
  Journal of Robotics Research}, vol.~39, no. 10-11, pp. 1326--1345, 2020.

\bibitem{chai2019multipath}
Y.~Chai, B.~Sapp, M.~Bansal, and D.~Anguelov, ``Multipath: Multiple
  probabilistic anchor trajectory hypotheses for behavior prediction,''
  \emph{arXiv preprint arXiv:1910.05449}, 2019.

\bibitem{rhinehart2019precog}
N.~Rhinehart, R.~McAllister, K.~Kitani, and S.~Levine, ``Precog: Prediction
  conditioned on goals in visual multi-agent settings,'' in \emph{Proceedings
  of the IEEE International Conference on Computer Vision}, 2019, pp.
  2821--2830.

\bibitem{zhao2020tnt}
H.~Zhao, J.~Gao, T.~Lan, C.~Sun, B.~Sapp, B.~Varadarajan, Y.~Shen, Y.~Shen,
  Y.~Chai, C.~Schmid \emph{et~al.}, ``Tnt: Target-driven trajectory
  prediction,'' \emph{arXiv preprint arXiv:2008.08294}, 2020.

\bibitem{hu2020scenario}
Y.~Hu, W.~Zhan, and M.~Tomizuka, ``Scenario-transferable semantic graph
  reasoning for interaction-aware probabilistic prediction,'' \emph{arXiv
  preprint arXiv:2004.03053}, 2020.

\bibitem{osher2004level}
S.~Osher, R.~Fedkiw, and K.~Piechor, ``Level set methods and dynamic implicit
  surfaces,'' \emph{Appl. Mech. Rev.}, vol.~57, no.~3, pp. B15--B15, 2004.

\bibitem{optimizedp}
\BIBentryALTinterwordspacing
M.~Bui, ``Optimized dynamic programming,'' 2020. [Online]. Available:
  \url{https://github.com/SFU-MARS/optimized\_dp}
\BIBentrySTDinterwordspacing

\bibitem{lai2019heterocl}
Y.-H. Lai, Y.~Chi, Y.~Hu, J.~Wang, C.~H. Yu, Y.~Zhou, J.~Cong, and Z.~Zhang,
  ``Heterocl: A multi-paradigm programming infrastructure for software-defined
  reconfigurable computing,'' \emph{Int'l Symp. on Field-Programmable Gate
  Arrays (FPGA)}, 2019.

\bibitem{interactiondataset}
W.~Zhan, L.~Sun, D.~Wang, H.~Shi, A.~Clausse, M.~Naumann, J.~K\"ummerle,
  H.~K\"onigshof, C.~Stiller, A.~de~La~Fortelle, and M.~Tomizuka,
  ``{INTERACTION} {Dataset}: {An} {INTERnational}, {Adversarial} and
  {Cooperative} {moTION} {Dataset} in {Interactive} {Driving} {Scenarios} with
  {Semantic} {Maps},'' \emph{arXiv:1910.03088 [cs, eess]}, 2019.

\bibitem{walambe2016optimal}
R.~Walambe, N.~Agarwal, S.~Kale, and V.~Joshi, ``Optimal trajectory generation
  for car-type mobile robot using spline interpolation,''
  \emph{IFAC-PapersOnLine}, vol.~49, no.~1, pp. 601--606, 2016.

\bibitem{jia2020ide}
X.~Jia, L.~Sun, M.~Tomizuka, and W.~Zhan, ``Ide-net: Interactive driving event
  and pattern extraction from human data,'' \emph{arXiv preprint
  arXiv:2011.02403}, 2020.

\bibitem{macqueen1967some}
J.~MacQueen \emph{et~al.}, ``Some methods for classification and analysis of
  multivariate observations,'' in \emph{Proceedings of the fifth Berkeley
  symposium on mathematical statistics and probability}, vol.~1, no.~14.\hskip
  1em plus 0.5em minus 0.4em\relax Oakland, CA, USA, 1967, pp. 281--297.

\bibitem{kong2015kinematic}
J.~Kong, M.~Pfeiffer, G.~Schildbach, and F.~Borrelli, ``Kinematic and dynamic
  vehicle models for autonomous driving control design,'' in \emph{2015 IEEE
  Intelligent Vehicles Symposium (IV)}.\hskip 1em plus 0.5em minus 0.4em\relax
  IEEE, 2015, pp. 1094--1099.

\bibitem{thrun2006stanley}
S.~Thrun, M.~Montemerlo, H.~Dahlkamp, D.~Stavens, A.~Aron, J.~Diebel, P.~Fong,
  J.~Gale, M.~Halpenny, G.~Hoffmann \emph{et~al.}, ``Stanley: The robot that
  won the darpa grand challenge,'' \emph{Journal of field Robotics}, vol.~23,
  no.~9, pp. 661--692, 2006.

\end{thebibliography}

\end{document}